\documentclass{article}

\usepackage[preprint]{neurips_2026}

\usepackage[utf8]{inputenc} 
\usepackage[T1]{fontenc}    
\usepackage{hyperref}       
\usepackage{url}            
\usepackage{booktabs}       
\usepackage{amsfonts}       
\usepackage{amssymb}
\usepackage{nicefrac}       
\usepackage{microtype}      
\usepackage{xcolor}         

\usepackage{graphicx}
\usepackage{multirow}
\usepackage{pifont}
\usepackage{amsmath}
\usepackage{algorithm}
\usepackage{algpseudocode}
\newcommand{\cmark}{\ding{51}}
\newcommand{\xmark}{\ding{55}}

\title{Parameter-free Adaptive Sparse Attention via Compression-Based Content Selection}

\author{ Debarshi Kundu \\ The Pennsylvania State University \\ \texttt{dqk5620@psu.edu} \AND Swaroop Ghosh \\ The Pennsylvania State University \\ \texttt{szg212@psu.edu} \And Vasant Honavar \\ The Pennsylvania State University \\ \texttt{vuh14@psu.edu} }

\begin{document}

\maketitle

\begin{abstract}
Data-adaptive sparse attention masks substantially outperform fixed patterns
(e.g., BigBird and Longformer) and can even exceed dense attention on long
sequences. Existing adaptive approaches---including SBM-Transformer,
Dynamic Mask Attention, and NSA---typically require additional learnable
parameters, custom gradient estimators, or specialized CUDA kernels.

We show that classical data compression provides an effective masking signal
with \textbf{no additional parameters}. By computing per-block gzip
compression ratios, we identify non-redundant content blocks and route
long-range attention selectively through them. Intuitively, blocks that
gzip cannot compress contain information not predictable from local
repetition, making them natural long-range attention targets. Because the
compression profile is input-dependent, the resulting sparse mask adapts
dynamically to content without learned parameters, auxiliary losses, or
custom kernels.

On PG-19 byte-level language modeling at 92M parameters with 8K context,
our method achieves 1.71 bits-per-byte (BPB), outperforming dense
attention (2.89), BigBird (2.34), Longformer (3.21), and a reimplemented
SBM-Transformer (3.38)---the only learned-mask baseline---by up to
1.67 BPB while adding no parameters. The advantage grows with sequence
length, with the gap over BigBird widening from 0.05 BPB at 4K context to
0.63 BPB at 8K, while convergence is 3.3$\times$ faster.
\end{abstract}

\section{Introduction}
\label{sec:intro}

Self-attention is the computational bottleneck of Transformers on long
sequences. Because attention scales quadratically with sequence length
$N$, processing 8K--64K token contexts requires either substantial
hardware or sparse attention mechanisms that restrict which positions each
token can attend to. The central challenge is therefore \emph{which
positions should be connected}---a question that has driven increasingly
sophisticated sparse attention methods.

\paragraph{Fixed patterns: simple but blind.}
Early sparse attention methods used predetermined connectivity patterns.
Longformer~\cite{beltagy2020longformer} employs a sliding window with
fixed global tokens, while BigBird~\cite{zaheer2020bigbird} augments this
with random long-range edges. These approaches are efficient and simple,
but their connectivity is \emph{independent of the input}: the same mask
is used whether the sequence is a legal contract or a children's story.
At shorter contexts (1K--4K tokens), this is often sufficient because the
fixed pattern covers enough useful connections by chance. At longer
contexts, however, the combinatorial space of useful connections grows
quadratically, and fixed patterns increasingly fail.

\paragraph{Learned masks: effective but complex.}
Recent work addresses this limitation by learning masks from the input.
Routing Transformer~\cite{roy2021routing} clusters queries using online
$k$-means and restricts attention within clusters. SBM-Transformer~\cite{cho2022sbm}
models attention as a stochastic block model with learnable memberships
and Bernoulli mask sampling via straight-through estimation. NSA~\cite{yuan2025nsa}
uses learned gating networks to select compressed token blocks, while
Dynamic Mask Attention trains per-head mask predictors using specialized
CUDA kernels. These methods consistently outperform fixed patterns, but
they introduce additional parameters, training complexity, custom
estimators, or specialized kernels.

\paragraph{Our insight: compression already provides the signal.}
We observe that the question ``which positions contain information worth
attending to?'' closely mirrors the objective solved by data compression.
If gzip compresses a text block to 90\% of its original size, the block
contains content not predictable from local repetition and is therefore a
natural long-range attention target. Conversely, blocks compressed to
30\% are largely repetitive. Compression ratio thus provides a direct
signal for identifying non-redundant content likely to benefit from
long-range attention.

We therefore propose \textbf{compression-guided sparse attention}: divide
the input into fixed-size blocks, compute per-block gzip compression
ratios, and route long-range attention only through high-ratio
(``literal'') blocks. The method adds zero learnable parameters, requires
no gradient estimators or custom kernels, and runs in $O(N)$ time using
standard-library compression. Because gzip ratios depend on input content,
the resulting sparse mask adapts naturally to each sequence.

\paragraph{Contributions.}
We make four contributions.
(i)~We introduce \textbf{compression-guided sparse attention}, the first
method to use classical data compression for adaptive attention-mask
construction, requiring zero additional parameters and no custom kernels.
(ii)~Through controlled ablations, we provide a clean
\textbf{component decomposition}: gzip-derived literal-to-literal
connections account for $-$0.43 BPB (68\% of the improvement over
BigBird), the local window for $-$0.12 BPB (19\%), and removing fixed
global tokens for $-$0.08 BPB (13\%).
(iii)~We show that \textbf{learned mask construction degrades at 8K
context} where zero-parameter compression succeeds (SBM-Transformer
3.38 BPB vs.\ ours 1.71), suggesting that external compression signals
can be more robust than gradient-based mask optimization at scale.
(iv)~We demonstrate that the advantage \textbf{grows with sequence
length}: the gap over BigBird widens from 0.05 BPB at 4K context to
0.63 BPB at 8K, while BigBird's delayed breakthrough shifts later as
context increases (step $\sim$4K at 4K, $\sim$7K at 8K).
\section{Related Work}
\label{sec:related}

\paragraph{Fixed-pattern sparse attention.}
Sparse Transformer~\cite{child2019sparse} introduced strided and local
attention patterns, showing that sparse attention can match dense
autoregressive modeling performance. Longformer~\cite{beltagy2020longformer}
uses a sliding window with fixed global tokens, while BigBird~\cite{zaheer2020bigbird}
adds random long-range edges and proves the resulting pattern is a
universal approximator of sequence functions. These methods are efficient
($O(N)$) and parameter-free, but their masks are
\emph{input-independent}: the same connectivity pattern is used regardless
of content. Our experiments show that this limitation becomes increasingly
costly as context length grows, with BigBird's convergence delayed at
larger sequence lengths (Table~\ref{tab:scaling}).

\paragraph{Learned adaptive sparse attention.}
To overcome the rigidity of fixed patterns, several methods learn masks
from the input. Routing Transformer~\cite{roy2021routing} clusters
queries and keys using online $k$-means and restricts attention within
clusters. SBM-Transformer~\cite{cho2022sbm} models connectivity as a
stochastic block model with learnable cluster embeddings and Bernoulli
mask sampling via straight-through estimation~\cite{bengio2013ste}.
NSA~\cite{yuan2025nsa} uses learned gating networks with hardware-aligned
CUDA kernels to select compressed token blocks, while Dynamic Mask
Attention trains per-head mask predictors with fused FlashAttention
kernels. SparseK Attention~\cite{lou2024sparsek} employs a differentiable
top-$k$ operator to select keys per query. These approaches consistently
outperform fixed patterns, but require additional parameters and training
complexity. In contrast, our method uses an \emph{external} compression
signal to construct adaptive masks with zero additional parameters and no
gradient flow through mask generation.

\paragraph{Compression in ML and long-context evaluation.}
Our work also connects to broader efforts using classical compression as
a structural signal in machine learning~\cite{jiang2023gzip,raff2024neuralncd}
and to methodology for evaluating long-context utilization~\cite{fang2025longppl,tay2021lra,rae2020compressive}.
We defer extended discussion to Appendix~\ref{app:extended_related} and
adopt bits-per-byte (BPB) on PG-19 as our primary metric, supplemented by
context-scaling analysis (Table~\ref{tab:context_scaling}).

\section{Method}
\label{sec:method}

\begin{figure}[!htb]
\centering
\includegraphics[width=\textwidth]{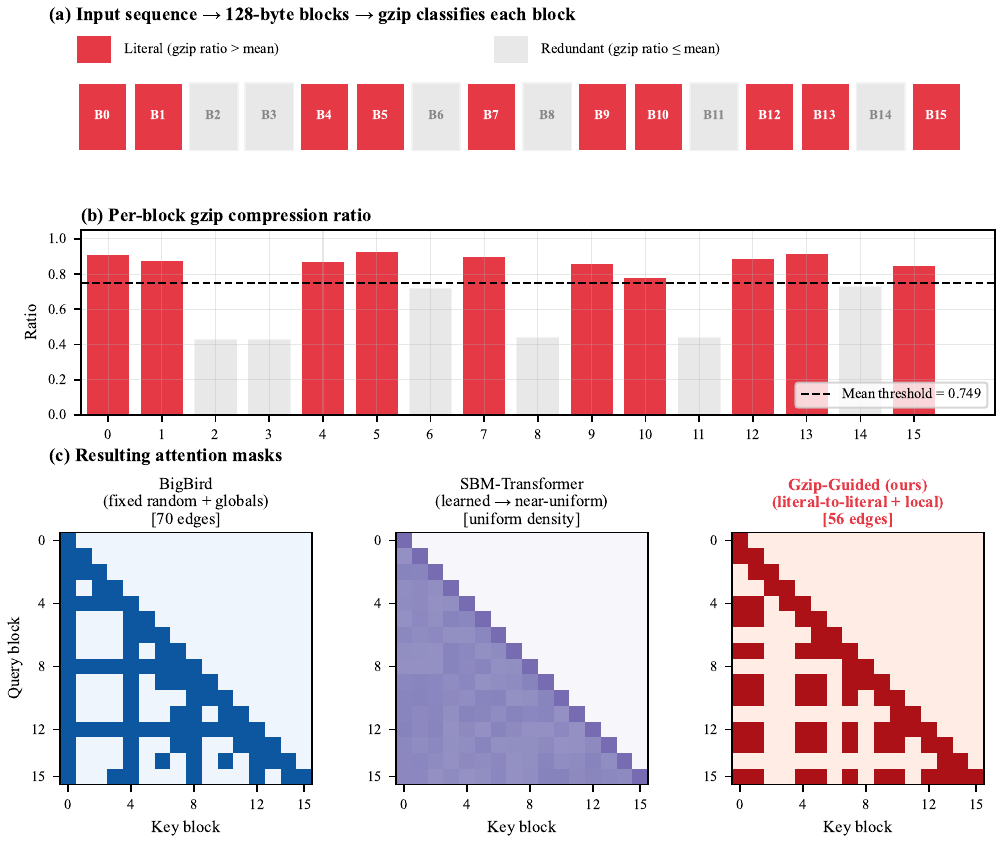}
\caption{\textbf{Method overview.} (a)~A sequence is divided into 128-byte
blocks. Gzip identifies \textcolor{red}{literal blocks} (high compression
ratio, content not predictable from within-block repetition) and
\textcolor{gray}{redundant blocks} (low ratio, repetitive or formulaic
content). (b)~Per-block compression ratios; blocks above the mean
threshold are literal. (c)~Resulting attention masks compared to BigBird
(fixed random + globals) and SBM-Transformer (learned, near-uniform). Our
mask connects literal blocks to each other, producing content-adaptive
sparse connectivity.}
\label{fig:overview}
\end{figure}

We construct sparse attention masks from classical data compression,
requiring zero learnable parameters. The method has three
components---block-level compression analysis (\S\ref{sec:compression}),
mask construction (\S\ref{sec:mask}), and integration with multi-head
attention (\S\ref{sec:integration})---followed by a complexity analysis
(\S\ref{sec:complexity}). Figure~\ref{fig:overview} illustrates the pipeline.

\subsection{Block-Level Compression Analysis}
\label{sec:compression}

Given an input sequence of $N$ bytes, we partition it into $B = N/b$ blocks
of $b$ bytes each ($b = 128$ throughout). For each block $i$, we
compute its \emph{compression ratio}:
\begin{equation}
r_i = \frac{|\texttt{gzip}(\mathbf{x}_{ib:(i+1)b})|}{b}
\label{eq:ratio}
\end{equation}
where $|\texttt{gzip}(\cdot)|$ denotes the compressed size in bytes using
gzip at compression level 1. The ratio $r_i \in (0, 1]$ measures
the information density of block $i$: high $r_i$ indicates content gzip
cannot compress---novel, non-redundant information---while low $r_i$
indicates predictable or repetitive patterns. Any general-purpose
lossless compressor would suffice; we use gzip~\cite{ziv1977lz77} for
its $O(N)$ runtime and standard-library availability.

We classify each block as \emph{literal} (informationally dense) or
\emph{redundant} using a parameter-free mean threshold:
\begin{equation}
\ell_i =
\begin{cases}
1 & \text{if } r_i > \overline{r} \\
0 & \text{otherwise}
\end{cases},
\qquad \overline{r} = \frac{1}{B}\sum_{j=1}^{B} r_j
\label{eq:literal}
\end{equation}
The mean threshold adapts naturally to each sequence and, unlike a
median, allows the literal fraction itself to vary with content.
When the ratio distribution is skewed toward low values (e.g., repetitive
prose), only a few blocks exceed the mean and the mask is sparse.
For denser technical content, the mask becomes correspondingly denser.
Empirically, the literal fraction varies from roughly 36\% to 67\%
across PG-19 sequences (Appendix~\ref{app:edges}), yielding a
3.5$\times$ range in long-range edge count without any sparsity
hyperparameter.

\subsection{Mask Construction}
\label{sec:mask}

We construct a block-level attention mask
$\mathbf{M} \in \{0,1\}^{B \times B}$ with three components:

\paragraph{Literal-to-literal connections.}
All pairs of literal blocks are connected for long-range attention:
$\mathbf{M}^{\text{lit}}_{ij} = 1$ iff $\ell_i = \ell_j = 1$ and
$|i-j| > w$, where $w$ is the local window size. Literal blocks thus
form an attention backbone among positions carrying non-redundant content.

\paragraph{Local sliding window.}
Adjacent blocks are always connected:
$\mathbf{M}^{\text{local}}_{ij} = 1$ iff $|i-j| \leq w$.
We use $w = 1$, providing 256 bytes of guaranteed local context for
redundant blocks predictable from nearby context.

\paragraph{No global tokens.}
Unlike BigBird~\cite{zaheer2020bigbird} and Longformer~\cite{beltagy2020longformer},
we use no fixed-position global tokens. Experiments consistently show
that global tokens degrade performance
(Section~\ref{sec:exp_ablation}), likely because they bias attention
toward positions selected by index rather than content.

The final block-level mask is
\begin{equation}
\mathbf{M} = \mathbf{M}^{\text{lit}} \cup \mathbf{M}^{\text{local}} \cup \mathbf{I}
\label{eq:full_mask}
\end{equation}
where $\mathbf{I}$ is the identity (self-attention). The block-level mask
is expanded to token resolution by repeating each entry across the
corresponding $b \times b$ token block, then intersected with a causal
mask to enforce autoregressive ordering. Because compression profiles
vary across inputs, $\mathbf{M}$ naturally adapts per sequence---more
complex inputs receive denser long-range connectivity.

\subsection{Integration with Multi-Head Attention}
\label{sec:integration}

We divide attention heads into three groups: 50\% \emph{local} heads
(use only $\mathbf{M}^{\text{local}} \cup \mathbf{I}$), 25\%
\emph{long-range} heads (use only
$\mathbf{M}^{\text{lit}} \cup \mathbf{I}$), and 25\% \emph{hybrid}
heads (use the full mask). All heads share the same mask across layers,
computed once per sequence and expanded to token resolution with a
single vectorized tensor operation.

\subsection{Complexity and Comparison to Learned Masks}
\label{sec:complexity}

\paragraph{Computational cost.}
Mask construction requires one gzip call per block:
$O(B) = O(N/b)$ operations, each compressing $b = 128$ bytes.
On modern hardware this takes $<1$ms for $N = 8192$, negligible
compared to attention itself. The resulting attention complexity depends
on the number of literal blocks $L$: each literal block attends to
$O(L)$ others, yielding $O(L^2 + Nw)$ total attention operations,
which is $O(N)$ when $L = O(\sqrt{N})$.

\paragraph{Block-to-token expansion.}
The block mask $\mathbf{M} \in \{0,1\}^{B \times B}$ is expanded to token
resolution by indexing entry $(i,j)$ from
$\mathbf{M}[\lfloor i/b \rfloor, \lfloor j/b \rfloor]$, implemented as a
single \texttt{expand}+\texttt{reshape} operation in PyTorch with no
Python loops.

\paragraph{Comparison to learned alternatives.}
A detailed comparison with adaptive sparse attention methods
(parameter counts, differentiability, kernel requirements, mask cost)
is given in Table~\ref{tab:method_comparison} of
Appendix~\ref{app:method_comp}. The key distinction is that our mask is
derived from an \emph{external} information-theoretic signal rather than
learned parameters. This eliminates the need for gradient flow through
mask construction---no straight-through estimators, Gumbel-softmax
relaxations, or custom backward passes. The compression ratio provides a
noise-free structural signal that scales to arbitrary sequence lengths
without additional training or architectural modifications.

\section{Experiments and Results}
\label{sec:experiments}

We organize the empirical study around five questions, each tested with a
controlled experiment. All experiments use the common setup in
Section~\ref{sec:common_setup}; experiment-specific changes are noted
below.

\subsection{Common Experimental Setup}
\label{sec:common_setup}

\textbf{Model.}~Byte-level autoregressive Transformer (ByteLM, 92M params):
768 hidden dim, 12 layers, 12 heads, 3072 FFN width, vocabulary 256, tied
input/output embeddings, gradient checkpointing, and bfloat16 mixed precision.
\textbf{Dataset.}~PG-19~\cite{rae2020compressive} (28,602 Project Gutenberg
books), encoded as raw UTF-8 bytes and chunked to 8192 bytes
($\approx$1500 words). A fixed random ordering (seed 42) ensures that all
methods within an experiment see identical data.
\textbf{Training.}~AdamW~\cite{loshchilov2019adamw} with learning rate
$3\times10^{-4}$, weight decay 0.1, $\beta=(0.9,0.95)$, 2000-step linear
warmup followed by cosine decay, 20K total steps, and effective batch size
16--20 across 4--5 H100 GPUs. All models use initialization seed 42.
\textbf{Evaluation.}~Bits-per-byte (BPB = cross-entropy$/\ln 2$) on the
PG-19 validation split, computed over 30 chunks at the training sequence
length.
\textbf{Mask construction.}~Each sequence is divided into 128-byte blocks;
we compute $r_i=|\texttt{gzip}(b_i)|/128$, mark blocks above the
per-sequence mean as ``literal'' (informationally dense), and connect all
literal-to-literal pairs for long-range attention, plus a $\pm1$-block
local window. No fixed-position global tokens are used.

\subsection{Experiment 1: How does compression-guided attention compare to existing methods?}
\label{sec:exp_main}

We compare seven attention methods using the same 92M-parameter model,
PG-19 data, sequence length 8192, and 20K training steps. Methods differ
only in mask construction:
\begin{itemize}
\setlength\itemsep{0pt}
\item \textbf{Dense:} Full $O(N^2)$ attention.
\item \textbf{Local Only:} $\pm1$-block sliding window (256 bytes); no globals or long-range edges.
\item \textbf{Longformer}~\cite{beltagy2020longformer}: $\pm4$-block window + global tokens every 8 blocks.
\item \textbf{BigBird}~\cite{zaheer2020bigbird}: Local + global tokens + 6 fixed random long-range edges per block.
\item \textbf{SBM-Transformer}~\cite{cho2022sbm}: Learned stochastic block model with per-head clusters, Bernoulli sampling, and STE; +247K parameters. Reimplemented from official code.\footnote{\url{https://github.com/sc782/SBM-Transformer}}
\item \textbf{Literal Pure} (ours): Gzip-identified literal blocks all-pairs only; no local or globals.
\item \textbf{Gzip-Guided} (ours): Literal-to-literal connections + local sliding window; no globals.
\end{itemize}

All methods share initialization, data order, optimizer, schedule, and
evaluation. For Literal Pure and Gzip-Guided, a fresh mask is constructed
per sequence from its compression profile; fixed-mask methods reuse the
same mask throughout training.

\begin{table}[!htb]
\centering
\caption{\textbf{Main results.} PG-19 language modeling at 92M parameters,
SEQ\_LEN=8192, 20K steps. All methods share architecture, initialization,
data ordering, and hyperparameters. Lower BPB is better.}
\label{tab:main}
\begin{tabular}{lcr}
\toprule
\textbf{Method} & \textbf{Extra Params} & \textbf{BPB} \\
\midrule
Longformer              & 0    & 3.21 \\
SBM-Transformer         & 247K & 3.38 \\
Dense Attention         & 0    & 2.89 \\
\midrule
Local Only              & 0    & 2.26 \\
BigBird                 & 0    & 2.34 \\
\midrule
Literal Pure (ours)     & 0    & 1.83 \\
\textbf{Gzip-Guided (ours)} & \textbf{0} & \textbf{1.71} \\
\bottomrule
\end{tabular}
\end{table}

Table~\ref{tab:main} shows a clear stratification by mask type.
Compression-guided methods achieve 1.71--1.83 BPB, substantially
outperforming fixed-pattern and learned-mask baselines. Gzip-Guided
achieves the best BPB (1.71) with zero extra parameters, improving over
Dense (2.89), BigBird (2.34), and Local Only (2.26) by 1.18, 0.63, and
0.55 BPB, respectively. SBM-Transformer, the only baseline that learns
the mask (+247K parameters and STE), performs worst at 3.38 BPB.

Even \emph{Literal Pure}, which uses only gzip-derived literal-to-literal
connections, reaches 1.83 BPB and outperforms every baseline. Thus,
compression-identified blocks form a strong long-range attention backbone
even without local windows or globals. A second pattern is that
fixed-position global tokens hurt: Local Only (2.26, no globals)
outperforms BigBird (2.34, with globals) despite BigBird's additional
random edges, suggesting that attention budget spent on content-agnostic
positions is wasted. A bar-chart visualization appears in
Figure~\ref{fig:method_comparison} (Appendix~\ref{app:supp_figs}).

\begin{figure}[!htb]
\centering
\includegraphics[width=\textwidth]{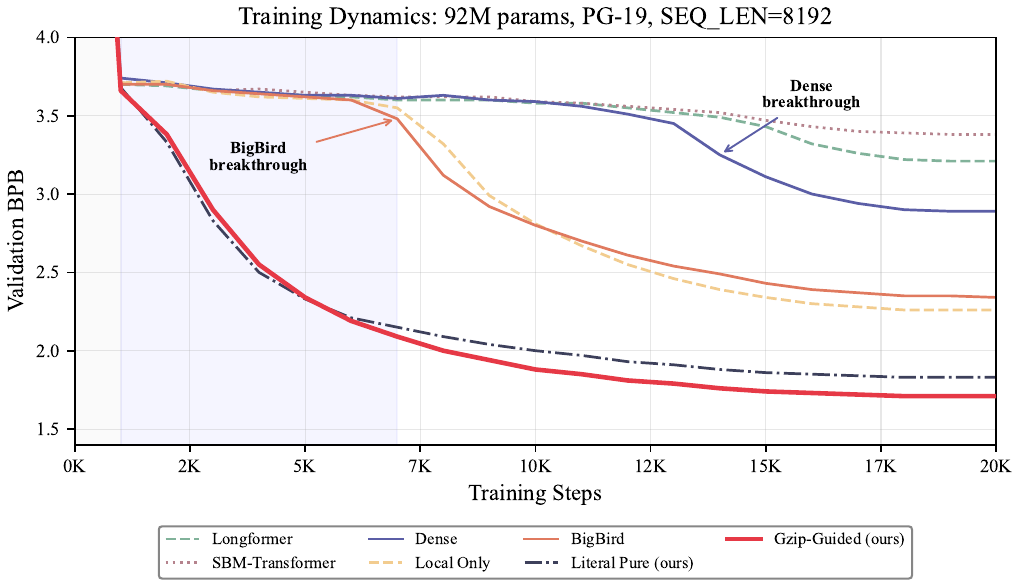}
\caption{\textbf{Convergence Dynamics.} Validation BPB during training.
Compression-guided methods (solid, bottom) diverge from fixed-mask
methods at step $\sim$1500 and descend continuously. BigBird and Local
exhibit delayed breakthroughs at steps 7K and 8K respectively.
SBM-Transformer drifts slowly downward without breakthrough despite
247K learned mask parameters.}
\label{fig:learning_curves}
\end{figure}

\paragraph{Convergence dynamics.}
Compression-guided and fixed-mask methods exhibit qualitatively different
learning curves (Figure~\ref{fig:learning_curves}).
All methods learn similarly for the first $\sim$1000 steps, after which
compression-guided methods accelerate sharply. Fixed-pattern methods
plateau for thousands of steps before delayed breakthroughs: BigBird at
$\sim$7K, Local at $\sim$8K, and Dense at $\sim$14K; SBM-Transformer
never breaks through. Gzip-Guided reaches 2.5 BPB in $\sim$4.2K steps,
whereas BigBird requires $\sim$14K steps (3.3$\times$ speedup), and
Dense, Longformer, and SBM do not reach this target within 20K steps.
Per-target step counts are in Appendix~\ref{app:convergence_table}.
\subsection{Experiment 2: Component ablation}
\label{sec:exp_ablation}

Our method differs from BigBird by (i) removing fixed global tokens,
(ii) replacing fixed random edges with gzip-derived literal-to-literal
edges, and (iii) adding a local sliding window. We isolate these effects
with controlled variants using the same 92M-parameter model, initialization,
data ordering, and schedule.

\begin{table}[!htb]
\centering
\caption{\textbf{Component ablation.} Each row modifies one component
relative to the previous. All results use 92M params, PG-19,
SEQ\_LEN=8192, and 20K steps.}
\label{tab:ablation}
\begin{tabular}{lccccc}
\toprule
\textbf{Configuration} & \textbf{Globals} & \textbf{Local} & \textbf{Long-range} & \textbf{BPB} & \textbf{$\Delta$ (\% Contr.)} \\
\midrule
BigBird (reference)              & \cmark & \cmark & Fixed random  & 2.34 & --- \\
$-$ globals (= Local Only)       & \xmark & \cmark & None          & 2.26 & $-$0.08 (13\%) \\
$+$ gzip literal (= Literal Pure)& \xmark & \xmark & Gzip-adaptive & 1.83 & $-$0.43 (68\%) \\
$+$ local window (= Gzip-Guided) & \xmark & \cmark & Gzip-adaptive & 1.71 & $-$0.12 (19\%) \\
\bottomrule
\end{tabular}

\footnotesize{Cumulative improvement from BigBird to Gzip-Guided: $-$0.63 BPB.
$\Delta$ values are relative to the previous row.}
\end{table}

The ablation reveals a clear hierarchy. \textbf{Removing global tokens
helps} ($-$0.08 BPB, 13\%): Local Only outperforms BigBird, indicating
that fixed-position globals bias attention toward index-selected rather
than content-selected positions. \textbf{Gzip-derived literal connections
dominate} ($-$0.43 BPB, 68\%): adding literal-to-literal edges drops BPB
from 2.26 to 1.83, showing that compression-derived edges form an
effective long-range routing backbone. \textbf{The local window adds
complementary benefit} ($-$0.12 BPB, 19\%): adding a $\pm1$-block window
to Literal Pure yields the full Gzip-Guided model at 1.71 BPB.

\paragraph{Long-range connections alone are nearly sufficient.}
Literal Pure, with only gzip-derived long-range edges and no local window,
already achieves 1.83 BPB, outperforming BigBird by 0.51 BPB and Local
Only by 0.43 BPB. Thus, the compression-identified graph is not merely a
supplement to local attention but a viable backbone aligned with the
input's information-theoretic structure.

\subsection{Experiment 3: How does the advantage scale with sequence length?}
\label{sec:exp_scaling}

We train BigBird and Gzip-Guided at 4096 and 8192 bytes using the same
92M-parameter architecture and 20K steps. At 4096 the sequence has 32
blocks; at 8192 it has 64 blocks, increasing the block-pair search space
from $32^2=1024$ to $64^2=4096$.

\begin{table}[!htb]
\centering
\caption{\textbf{Sequence length scaling.} Gzip-Guided outperforms BigBird
at both sequence lengths. The gap widens from 0.05 BPB at 4K to 0.63 BPB
at 8K as the attention search space quadruples.}
\label{tab:scaling}
\begin{tabular}{lccccc}
\toprule
\textbf{SEQ\_LEN} & \textbf{Blocks} & \textbf{BigBird} & \textbf{Gzip-Guided} & \textbf{Gap} & \textbf{Speedup} \\
\midrule
4096   & 32  & 1.64 & 1.59 & 0.05 & 1.8$\times$ \\
8192   & 64  & 2.34 & 1.71 & 0.63 & 3.3$\times$ \\
\bottomrule
\end{tabular}
\end{table}

At 4K, the gap is only 0.05 BPB: fixed random masks work adequately when
the search space is small. At 8K, the gap widens to 0.63 BPB as the
quadrupled search space makes random patterns less likely to cover useful
long-range connections. BigBird's breakthrough is delayed from
$\sim$4K to $\sim$7K steps and converges to a worse final BPB. This trend
matches a simple counting argument: as sequences grow, useful long-range
dependencies become a smaller fraction of all $O(N^2)$ possible pairs, so
random edges become poorer approximations of useful routes. Compression
targets non-redundant positions directly and therefore scales more
gracefully. Learning curves appear in Figure~\ref{fig:scaling_curves}
(Appendix~\ref{app:supp_figs}).

\subsection{Experiment 4: Can learned masks compete with compression?}
\label{sec:exp_sbm}

SBM-Transformer~\cite{cho2022sbm} is the most directly comparable learned
adaptive mask method. We reimplement it from the official code,\footnote{\url{https://github.com/sc782/SBM-Transformer/tree/main/code}}
adapting it to autoregressive byte-level modeling while preserving key
details: \texttt{kaiming\_normal\_} cluster initialization, shared
two-layer ReLU MLP node projections, flattened softmax over $k^2$ block
entries, the custom STE (\texttt{SampleGraphSparseGraph}) using
\texttt{hardtanh}$(A\cdot g)$ in the backward pass, and the mask pipeline
(softmax $\to$ binary mask multiply $\to$ $L_1$ renormalization). We use
8 clusters per head and add a causal mask. The reimplementation adds 247K
parameters (0.27\% of the base model).

\begin{table}[!htb]
\centering
\caption{\textbf{Zero-parameter vs.\ learned masks.} SBM-Transformer adds
247K parameters for adaptive mask generation. Our method uses gzip---no
parameters, no STE, and no custom kernels.}
\label{tab:vs_learned}
\begin{tabular}{lcccr}
\toprule
\textbf{Method} & \textbf{Extra Params} & \textbf{Needs STE} & \textbf{Custom CUDA} & \textbf{BPB} \\
\midrule
SBM-Transformer  & 247K & Yes & No  & 3.38 \\
Gzip-Guided      & 0    & No  & No  & 1.71 \\
\midrule
\multicolumn{4}{l}{$\Delta$ (learned $-$ compression)} & +1.67 \\
\bottomrule
\end{tabular}
\end{table}

SBM-Transformer achieves 3.38 BPB, worse than dense attention (2.89) and
all sparse baselines.

\paragraph{Analysis: why does SBM fail at 8K?}
A likely failure mode is high-variance stochastic optimization. Bernoulli
sampling over $N^2=67$M potential edges at $N=8192$ requires the STE to
propagate useful signal through a binary stochastic mask that changes
each forward pass; at 8K, the signal-to-noise ratio appears insufficient
for cluster embeddings to learn useful structure. Its trajectory
(Table~\ref{tab:trajectories}) shows slow monotonic descent without the
breakthrough seen when effective patterns emerge.

This contrasts with the original SBM-Transformer results on Long Range
Arena~\cite{tay2021lra}, where sequences are 1K--4K tokens. At those
lengths, the search space ($N^2\leq16$M) is more manageable for STE.
Our results suggest that learned masks face a scaling challenge precisely
where compression provides a noise-free, parameter-free structural signal.

\subsection{Experiment 5: How do the methods utilize context?}
\label{sec:exp_context}

To test whether BPB gains reflect better long-range context use rather
than only better local representations, we evaluate trained models at
multiple context budgets. For budget $C$, we keep the full 8192-byte input
but zero out attention-mask columns corresponding to blocks before
position $(N-C)$. Each method uses its own mask type to avoid mask-switching
artifacts. We measure BPB on the last 512 bytes, which have exactly $C$
bytes of preceding context. This is clean for within-method comparisons;
cross-method deltas should be interpreted cautiously because each model is
optimized for its own mask. We exclude SBM-Transformer because its
stochastic Bernoulli mask is sensitive to train/eval-mode details, and
report only its training-time validation BPB in Table~\ref{tab:main}.

\begin{table}[!htb]
\centering
\caption{\textbf{Context utilization.} BPB on the last 512 bytes as
available context varies. Each method uses its own mask with early blocks
progressively blocked. $\Delta$ = improvement from 512 to 8192 bytes.}
\label{tab:context_scaling}
\begin{tabular}{lcccc}
\toprule
\textbf{Context} & \textbf{LF} & \textbf{BB} & \textbf{Local} & \textbf{Gzip} \\
\midrule
512   & 4.02 & 2.48 & 2.27 & 1.89 \\
1024  & 3.86 & 2.30 & 2.17 & 1.78 \\
2048  & 3.83 & 2.28 & 2.15 & 1.75 \\
4096  & 3.67 & 2.27 & 2.14 & 1.71 \\
8192  & 3.25 & 2.26 & 2.14 & 1.68 \\
\midrule
$\Delta$ & 0.77 & 0.22 & 0.13 & \textbf{0.21} \\
\bottomrule
\end{tabular}
\end{table}

Gzip-Guided achieves the lowest absolute BPB at every context length.
With only 512 bytes of context, it reaches 1.89 BPB---better than BigBird
with full 8192-byte context (2.26). This indicates that
compression-derived masks improve representation quality across context
scales, not only at long contexts. Within-method trends show that
Gzip-Guided extracts comparable long-context gains to BigBird
(0.21 vs.\ 0.22 BPB from 512 to 8192) while operating from a much lower
starting point, indicating more effective routing throughout the context
range.

\section{Summary and Discussion}
\label{sec:discussion}

\label{sec:conclusion}
\noindent\textbf{Summary.}
We introduced compression-guided sparse attention, which constructs
per-sequence attention masks from gzip compression ratios with zero
additional parameters: blocks gzip cannot compress carry non-redundant
information and become long-range attention targets; blocks gzip can
compress are handled by a short local window. On PG-19 at 92M
parameters and 8K context, our method achieves 1.71 BPB, outperforming
dense attention (2.89), BigBird (2.34), Local Only (2.26), and a
reimplemented SBM-Transformer (3.38). Compression-derived
literal-to-literal connections alone (no local window, no globals)
already reach 1.83 BPB, and the gap over fixed-pattern methods widens
with sequence length (0.05 at 4K to 0.63 at 8K).

More broadly, the result suggests classical compression algorithms
still have much to offer modern deep learning: when a problem can be
cast in terms of identifying redundant content, decades of engineering
on practical compressors provides parameter-free, deployment-ready
answers. Promising directions include integration with block-sparse
kernels for wall-clock speedups, scaling to billion-parameter models
with subword tokenization, and exploring stronger compressors (brotli,
zstd, neural compressors) that may yield sharper signals.

\label{sec:limitations}
\noindent\textbf{Limitations.}
We briefly note the principal limitations that qualify our findings;
each is discussed in full in Appendix~\ref{app:limitations}. Our
experiments use a single 92M-parameter model trained for 20K steps on
a single dataset (PG-19), with byte-level rather than subword
tokenization, and the only compressor we evaluate is gzip at level 1.
Our implementation materializes the full $N \times N$ attention matrix
and applies the sparse mask via element-wise multiplication, providing
no wall-clock speedup over dense attention---the sparsity is logical
rather than computational. Our SBM-Transformer baseline is a
reimplementation rather than the original authors' code on their
original benchmark, so its poor performance in our setting may reflect
domain mismatch with the encoder-based 1K--4K classification regime
for which it was designed. Finally, we report only PG-19 BPB; downstream
task transfer remains untested.

\noindent\textbf{Discussion.} The experimental results presented above raise the question as to why compression length is associated with attention.
The empirical pattern---gzip-guided attention reaching 1.71 BPB
vs.\ 2.34 for BigBird and 2.89 for dense---admits a simple intuition.
Self-attention solves a routing problem at every layer: given a query,
which keys carry information that reduces its predictive uncertainty?
For natural language, the keys that matter are those carrying novel
content, not those echoing predictable patterns already inferable from
nearby context. Gzip's compression ratio is a cheap proxy for this
distinction: blocks that compress well contain repetitive or formulaic
material, while blocks that do not compress contain content with low
internal redundancy. By connecting the latter to each other, we
construct an attention graph that preferentially links positions
carrying non-redundant content, while letting the local window handle
predictable material.

This intuition is supported by Literal Pure's strong performance
(1.83 BPB with \emph{only} literal-to-literal connections, no local window):
the compression-identified blocks form a sufficient backbone for effective
language modeling, suggesting alignment between gzip's notion of
non-redundancy and the model's need for long-range dependencies.
Appendix~\ref{app:extended_discussion} extends this discussion to
(i)~why an external compression signal outperforms a learned one in
our setting, (ii)~implications for sequence-length scaling, and
(iii)~the choice of compressor.



\bibliographystyle{plainnat}
\bibliography{references}


\appendix

\section{Hyperparameters and Compute}
\label{app:hyperparams}

\subsection{Model Architecture}

Table~\ref{tab:arch} details the model configuration used in all experiments
unless otherwise noted.

\begin{table}[!htb]
\centering
\caption{ByteLM architecture (medium configuration).}
\label{tab:arch}
\begin{tabular}{ll}
\toprule
\textbf{Hyperparameter} & \textbf{Value} \\
\midrule
Vocabulary size          & 256 (byte-level) \\
Model dimension ($d$)    & 768 \\
Attention heads ($H$)    & 12 \\
Head dimension ($d_k$)   & 64 \\
Layers ($L$)             & 12 \\
FFN dimension            & 3072 \\
Activation               & GELU \\
Dropout                  & 0.1 \\
Positional encoding      & Learned (max 16384) \\
Embedding tying          & Input = Output \\
Total parameters         & 91,544,064 \\
\bottomrule
\end{tabular}
\end{table}

At SEQ\_LEN=4096, the positional embedding table is smaller,
yielding 88,398,336 parameters. The SBM-Transformer adds 173,568
parameters (0.19\% overhead): 6,144 per layer for cluster embeddings
($H \times K \times d_k = 12 \times 8 \times 64$) and 8,320 per layer
for the shared projection MLP ($2 \times (d_k^2 + d_k) = 2 \times 4160$).

\subsection{Training Configuration}

\begin{table}[!htb]
\centering
\caption{Training hyperparameters.}
\label{tab:training_config}
\begin{tabular}{ll}
\toprule
\textbf{Hyperparameter} & \textbf{Value} \\
\midrule
Optimizer                & AdamW \\
Learning rate            & $3 \times 10^{-4}$ \\
Weight decay             & 0.1 \\
$\beta_1, \beta_2$       & 0.9, 0.95 \\
$\epsilon$               & $10^{-8}$ \\
LR schedule              & Linear warmup + cosine decay \\
Warmup steps             & 2000 \\
Training steps           & 20,000 \\
Batch size per GPU       & 1 \\
Gradient accumulation    & 4 \\
GPUs                     & 4--5 $\times$ NVIDIA H100 SXM \\
Effective batch size     & 16--20 sequences \\
Precision                & bfloat16 mixed precision \\
Gradient checkpointing   & Enabled (disabled for SBM) \\
Random seed              & 42 (model init + data order) \\
\bottomrule
\end{tabular}
\end{table}

\subsection{Mask Construction Parameters}

\begin{table}[!htb]
\centering
\caption{Sparse attention mask hyperparameters.}
\label{tab:mask_config}
\begin{tabular}{ll}
\toprule
\textbf{Parameter} & \textbf{Value} \\
\midrule
Block size ($b$)                    & 128 bytes \\
Local window ($w$)                  & $\pm 1$ block (256 bytes) \\
Gzip compression level              & 1 (fastest) \\
Literality threshold                & Mean compression ratio per sequence \\
Global tokens                       & None (removed in all our variants) \\
BigBird random edges ($K$)           & 6 per block \\
Longformer window                    & $\pm 4$ blocks \\
Longformer global stride             & Every 8th block \\
SBM clusters ($k$)                   & 8 per head \\
SBM edge exploration ($\delta$)      & 0.01 \\
\midrule
Head allocation: local               & 50\% of heads (6/12) \\
Head allocation: long-range          & 25\% of heads (3/12) \\
Head allocation: hybrid              & 25\% of heads (3/12) \\
\bottomrule
\end{tabular}
\end{table}

\subsection{Compute Budget}

\begin{table}[!htb]
\centering
\caption{Approximate compute costs per experiment.}
\label{tab:compute}
\begin{tabular}{llcc}
\toprule
\textbf{Experiment} & \textbf{Configuration} & \textbf{GPU-hours} & \textbf{Hardware} \\
\midrule
Main (8K, 20K steps) & 7 methods, each on 4 GPUs     & $7 \times 5 = 35$ & 4$\times$H100 \\
Scaling (4K, 20K)    & 2 methods, sequential         & 5                 & 4$\times$H100 \\
Evaluation suite     & 7 models $\times$ 3 evals     & 11                & 1$\times$H100 \\
\midrule
\textbf{Total}       &                               & $\sim$\textbf{51} & \\
\bottomrule
\end{tabular}
\end{table}

All experiments were run on RunPod cloud instances with NVIDIA H100 SXM
(80GB) GPUs. Training used PyTorch 2.x with DistributedDataParallel
across GPUs within each node. A shared network volume stored data and
checkpoints, enabling parallel execution of different methods on
separate pods.

\section{Mask Construction Pseudocode}
\label{app:pseudocode}

Algorithm~\ref{alg:mask} presents the complete mask construction procedure.

\begin{algorithm}[!htb]
\caption{Compression-Guided Sparse Attention Mask}
\label{alg:mask}
\begin{algorithmic}[1]
\Require Sequence $\mathbf{x} \in \{0,\ldots,255\}^N$, block size $b$, window $w$
\Ensure Token-level mask $\mathbf{M}' \in \{0,1\}^{N \times N}$, where $B = N/b$

\Statex
\Statex \textit{// Step 1: Per-block compression ratios \hfill [$O(B)$ gzip calls]}
\For{$i = 0$ to $B-1$}
    \State $r_i \gets |\texttt{gzip}(\mathbf{x}[ib:(i{+}1)b])| \;/\; b$
\EndFor

\Statex
\Statex \textit{// Step 2: Classify blocks \hfill [parameter-free]}
\State $\tau \gets \texttt{mean}(r_0, \ldots, r_{B-1})$
\For{$i = 0$ to $B-1$}
    \State $\ell_i \gets \mathbf{1}[r_i > \tau]$
\EndFor

\Statex
\Statex \textit{// Step 3: Build block-level mask \hfill [3 components]}
\State $\mathbf{M} \gets \mathbf{I}_B$ \Comment{diagonal (self-attn)}
\For{$i,j \in \{0,\ldots,B{-}1\}^2$}
    \If{$|i{-}j| \leq w$} $\mathbf{M}_{ij} \gets 1$ \Comment{local}
    \EndIf
    \If{$\ell_i{=}1$ \textbf{and} $\ell_j{=}1$ \textbf{and} $|i{-}j|{>}w$}
        \State $\mathbf{M}_{ij} \gets 1$ \Comment{literal}
    \EndIf
\EndFor

\Statex
\Statex \textit{// Step 4: Expand to tokens + causal masking \hfill [vectorized]}
\State $\mathbf{M}' \gets \texttt{expand}(\mathbf{M}, b)$ \Comment{$B{\times}B \to N{\times}N$}
\State $\mathbf{M}' \gets \mathbf{M}' \;\texttt{AND}\; \texttt{NOT}\;\texttt{triu}(\mathbf{1}, 1)$ \Comment{enforce causality}
\end{algorithmic}
\end{algorithm}

\noindent
Total cost is dominated by $B$ gzip calls on $b$-byte blocks ($<$1ms for
$N=8192$, $b=128$).

\paragraph{Head-specific mask distribution.}
The block mask $\mathbf{M}$ is split across attention heads:
\begin{itemize}
\setlength\itemsep{0.1em}
\item \textbf{Local heads} (heads 0--5): $\mathbf{M}^{\text{local}} \cup \mathbf{I}$ only.
\item \textbf{Long-range heads} (heads 6--8): $\mathbf{M}^{\text{lit}} \cup \mathbf{I}$ only.
\item \textbf{Hybrid heads} (heads 9--11): full $\mathbf{M}$.
\end{itemize}
This allocation is identical across all 12 layers and fixed throughout training.

\paragraph{Block-to-token expansion.}
Implemented as a single vectorized operation in PyTorch:

\noindent
\texttt{token\_mask = block\_mask.unsqueeze(-1).unsqueeze(-1)}\\
\texttt{\phantom{token\_mask = }.expand(B, B, b, b).reshape(N, N)}

\noindent
No Python loops. Cost is negligible relative to attention.

\section{SBM-Transformer Reimplementation Details}
\label{app:sbm}

We reimplemented SBM-Transformer following the official code at
\texttt{github.com/sc782/SBM-Transformer}. This section documents the
key implementation decisions.

\subsection{Components Preserved from Official Code}

\begin{enumerate}
\setlength\itemsep{0.2em}
\item \textbf{Cluster embeddings:} Per-head parameters
      $\mathbf{C} \in \mathbb{R}^{H \times K \times d_k}$ initialized with
      \texttt{kaiming\_normal\_} (matching the official code).

\item \textbf{Node projection:} A shared (not per-head)
      \texttt{nn.Sequential(Linear($d_k$,$d_k$), ReLU, Linear($d_k$,$d_k$))}
      applied to both queries and keys.

\item \textbf{Block matrix:} $\hat{\mathbf{S}} = \text{softmax}(\mathbf{C}\mathbf{C}^\top)$
      with softmax over the \emph{flattened} $K^2$ entries (matching the
      official \texttt{.reshape(num\_head, num\_clusters**2)}).

\item \textbf{Memberships:}
      $\hat{\mathbf{Q}} = \sigma(\text{MLP}(\mathbf{Q}) \cdot \mathbf{C}^\top)$,
      $\hat{\mathbf{K}} = \sigma(\text{MLP}(\mathbf{K}) \cdot \mathbf{C}^\top)$.

\item \textbf{Edge probabilities:}
      $\mathbf{A} = \hat{\mathbf{Q}} \hat{\mathbf{S}} \hat{\mathbf{K}}^\top$.

\item \textbf{Custom STE} (\texttt{SampleGraphSparseGraph}):
      Forward: $\mathbf{G} = \text{Bernoulli}(\text{clamp}(\mathbf{A}+0.01, 0, 1))$.
      Backward: $\nabla_{\mathbf{A}} = \text{hardtanh}(\mathbf{G} \odot \nabla_{\text{out}})$.

\item \textbf{Mask application:}
      $\text{attn} = L_1\text{-normalize}(\text{softmax}(\mathbf{QK}^\top/\sqrt{d_k}) \odot \mathbf{G})$
      (matching official \texttt{F.normalize(...*graph, p=1, dim=-1)}).
\end{enumerate}

\subsection{Adaptations for Our Setting}

\begin{enumerate}
\setlength\itemsep{0.2em}
\item \textbf{Causal mask:} Added standard upper-triangular masking to
      attention logits before softmax for autoregressive training.

\item \textbf{Clusters:} $K = 8$ (the argparse default in the official
      \texttt{run\_tasks.py}).

\item \textbf{Gradient checkpointing:} Disabled for SBM. The official
      code sets \texttt{attention\_grad\_checkpointing: False} for SBM,
      and we found empirically that Bernoulli resampling during checkpoint
      recomputation produces inconsistent gradients.

\item \textbf{NaN handling:} Added \texttt{nan\_to\_num} after $L_1$
      normalization. At 8K context, the intersection of Bernoulli sampling
      and causal masking occasionally produces all-zero rows.
\end{enumerate}

\subsection{Parameter Count}

\begin{table}[!htb]
\centering
\caption{SBM-Transformer additional parameters per layer.}
\label{tab:sbm_params}
\begin{tabular}{lrl}
\toprule
\textbf{Component} & \textbf{Params} & \textbf{Formula} \\
\midrule
Cluster embeddings   & 6,144  & $H \times K \times d_k = 12 \times 8 \times 64$ \\
MLP weight 1         & 4,096  & $d_k \times d_k$ \\
MLP bias 1           & 64     & $d_k$ \\
MLP weight 2         & 4,096  & $d_k \times d_k$ \\
MLP bias 2           & 64     & $d_k$ \\
\midrule
\textbf{Per layer}   & \textbf{14,464} & \\
\textbf{12 layers}   & \textbf{173,568} & (0.19\% of base model) \\
\bottomrule
\end{tabular}
\end{table}

\section{Full Training Trajectories}
\label{app:trajectories}

Table~\ref{tab:trajectories} provides the complete validation BPB at
every evaluation checkpoint for all seven methods.

\begin{table}[!htb]
\centering
\caption{Validation BPB at each evaluation checkpoint
(92M params, PG-19, SEQ\_LEN=8192, 20K steps).
BB = BigBird, LF = Longformer, Lit-P = Literal Pure.}
\label{tab:trajectories}
\footnotesize
\begin{tabular}{rccccccc}
\toprule
\textbf{Step} & \textbf{Dense} & \textbf{BB} & \textbf{LF} & \textbf{SBM} & \textbf{Local} & \textbf{Lit-P} & \textbf{Gzip} \\
\midrule
1K   & 3.74  & 3.70 & 3.70 & 3.71  & 3.71 & 3.68 & 3.66 \\
2K   & 3.71  & 3.70 & 3.69 & 3.70  & 3.72 & 3.33 & 3.38 \\
3K   & 3.67  & 3.66 & 3.66 & 3.66  & 3.65 & 2.83 & 2.90 \\
4K   & 3.65  & 3.64 & 3.64 & 3.67  & 3.62 & 2.50 & 2.55 \\
5K   & 3.63  & 3.62 & 3.63 & 3.65  & 3.61 & 2.33 & 2.34 \\
6K   & 3.63  & 3.60 & 3.62 & 3.63  & 3.60 & 2.21 & 2.19 \\
7K   & 3.61  & 3.48 & 3.60 & 3.62  & 3.55 & 2.15 & 2.09 \\
8K   & 3.63  & 3.12 & 3.60 & 3.62  & 3.32 & 2.09 & 2.00 \\
9K   & 3.60  & 2.92 & 3.60 & 3.62  & 2.99 & 2.04 & 1.94 \\
10K  & 3.59  & 2.80 & 3.58 & 3.59  & 2.81 & 2.00 & 1.88 \\
11K  & 3.56  & 2.70 & 3.58 & 3.58  & 2.67 & 1.97 & 1.85 \\
12K  & 3.51  & 2.61 & 3.55 & 3.56  & 2.55 & 1.93 & 1.81 \\
13K  & 3.45  & 2.54 & 3.52 & 3.54  & 2.46 & 1.91 & 1.79 \\
14K  & 3.25  & 2.49 & 3.49 & 3.52  & 2.39 & 1.88 & 1.76 \\
15K  & 3.11  & 2.43 & 3.43 & 3.47  & 2.34 & 1.86 & 1.74 \\
16K  & 3.00  & 2.39 & 3.32 & 3.43  & 2.30 & 1.85 & 1.73 \\
17K  & 2.94  & 2.37 & 3.26 & 3.40  & 2.28 & 1.84 & 1.72 \\
18K  & 2.90  & 2.35 & 3.22 & 3.39  & 2.26 & 1.83 & 1.71 \\
19K  & 2.89  & 2.35 & 3.21 & 3.38  & 2.26 & 1.83 & 1.71 \\
20K  & 2.89  & 2.34 & 3.21 & 3.38  & 2.26 & 1.83 & 1.71 \\
\bottomrule
\end{tabular}
\normalsize
\end{table}

Three training regimes are visible:
\begin{itemize}
\setlength\itemsep{0.1em}
\item \textbf{Steps 0--1K:} All methods learn identically (basic byte statistics).
\item \textbf{Steps 1K--7K:} Compression-guided methods accelerate sharply;
      fixed-mask methods plateau. At step 5K, Gzip-Guided is at 2.34 BPB
      while BigBird remains at 3.62.
\item \textbf{Steps 7K--20K:} Fixed-pattern methods undergo delayed
      breakthroughs (BigBird $\sim$7K, Local $\sim$8K, Dense $\sim$14K).
      The gap narrows but remains substantial (0.63 BPB for BigBird, 1.18 for Dense).
      SBM-Transformer drifts downward without breakthrough.
\end{itemize}

\section{Sequence Length Scaling: Detailed Results}
\label{app:scale_detail}

\subsection{SEQ\_LEN=4096}

\begin{table}[!htb]
\centering
\caption{Training trajectories at SEQ\_LEN=4096 (92M params, 20K steps).}
\label{tab:traj_4k}
\begin{tabular}{rcc}
\toprule
\textbf{Step} & \textbf{BigBird} & \textbf{Gzip-Guided} \\
\midrule
1K   & 3.64 & 3.58 \\
2K   & 3.61 & 2.91 \\
3K   & 3.59 & 2.37 \\
4K   & 3.45 & 2.12 \\
5K   & 2.85 & 1.98 \\
6K   & 2.46 & 1.88 \\
7K   & 2.22 & 1.82 \\
8K   & 2.07 & 1.78 \\
9K   & 1.96 & 1.75 \\
10K  & 1.86 & 1.72 \\
15K  & 1.68 & 1.62 \\
20K  & 1.64 & 1.59 \\
\bottomrule
\end{tabular}
\end{table}

At 4K, BigBird's breakthrough occurs at step $\sim$4K (vs.\ $\sim$7K at 8K)
and the final gap is only 0.05 BPB (vs.\ 0.63 at 8K). The smaller attention
search space ($32^2 = 1024$ vs.\ $64^2 = 4096$ block pairs) allows fixed
patterns to discover useful routes more quickly.

\begin{figure}[!htb]
\centering
\includegraphics[width=0.85\textwidth]{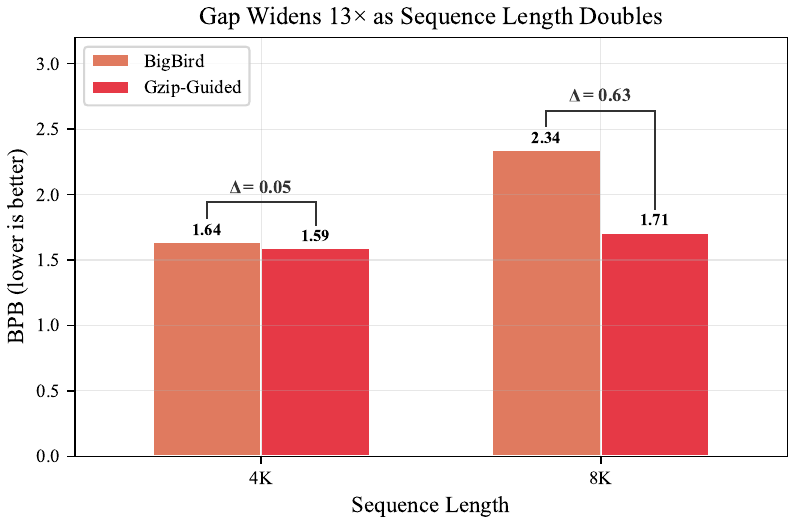}
\caption{Left: BPB comparison at 4K and 8K context. Right: the gap
between BigBird and Gzip-Guided at each sequence length.}
\label{fig:scaling_gap}
\end{figure}

\section{Context Scaling: Complete Results}
\label{app:context_full}

Table~\ref{tab:context_full} extends Table~\ref{tab:context_scaling} with
the Dense model and additional context lengths.

\begin{table}[!htb]
\centering
\caption{Complete context scaling results. BPB on last 512 bytes with
varying context budget. Each method uses its own mask type.}
\label{tab:context_full}
\footnotesize
\begin{tabular}{lccccc}
\toprule
\textbf{Ctx} & \textbf{Dense} & \textbf{LF} & \textbf{BB} & \textbf{Local} & \textbf{Gzip} \\
\midrule
256   & 4.14 & 4.00 & 3.63 & 3.27 & 5.23 \\
512   & 4.53 & 4.02 & 2.48 & 2.27 & 1.89 \\
1024  & 4.41 & 3.86 & 2.30 & 2.17 & 1.78 \\
2048  & 4.25 & 3.83 & 2.28 & 2.15 & 1.75 \\
4096  & 4.02 & 3.67 & 2.27 & 2.14 & 1.71 \\
8192  & 2.96 & 3.25 & 2.26 & 2.14 & 1.68 \\
\bottomrule
\end{tabular}
\normalsize
\end{table}

\paragraph{Dense model artifact.}
The Dense model shows non-monotonic behavior: BPB \emph{increases}
from 256 to 512 bytes (4.14$\to$4.53). This occurs because the Dense
model was trained without any block-level masking; introducing
column-blocking at evaluation disrupts its learned attention patterns.
Dense context-scaling numbers should be interpreted with caution for
cross-method comparison. The main table value (2.89 BPB) is computed
without any blocking and remains valid.

\paragraph{Short-context degradation in Gzip-Guided.}
Gzip-Guided performs poorly at 256 bytes (5.23 BPB) because almost no
sparse connections exist in such a small window (only 2 blocks).
Performance recovers sharply at 512 bytes (4 blocks), where the local
window and initial literal connections become available. This confirms
that the model's representations are structurally dependent on its
sparse attention pathways.

\section{Edge Count Distribution}
\label{app:edges}

The Gzip-Guided method produces a varying number of long-range edges per
sequence, reflecting content diversity in PG-19.

\begin{table}[!htb]
\centering
\caption{Edge count statistics for Gzip-Guided at SEQ\_LEN=8192
(computed over 1000 training sequences).}
\label{tab:edge_stats}
\begin{tabular}{ll}
\toprule
\textbf{Statistic} & \textbf{Value} \\
\midrule
Minimum edges   & 256 \\
Maximum edges   & 904 \\
Mean            & 593 \\
Median          & 621 \\
Std.\ dev.      & 135 \\
Range ratio     & 3.5$\times$ \\
\bottomrule
\end{tabular}
\end{table}

For comparison, BigBird always produces exactly 309 edges per sequence
(fixed by construction), regardless of input content.

The adaptive density reflects content structure: repetitive passages
(fairy tales with repeated phrases, formulaic letter openings) produce
fewer literal blocks (gzip compresses such content well) and thus fewer
edges. Dense novel content (technical descriptions, dialogue with unique
vocabulary) produces more literal blocks and denser connectivity. This
automatic compute allocation---more attention budget for less-redundant
inputs---is a practical consequence of the compression-based approach.

\begin{figure}[!htb]
\centering
\includegraphics[width=0.85\textwidth]{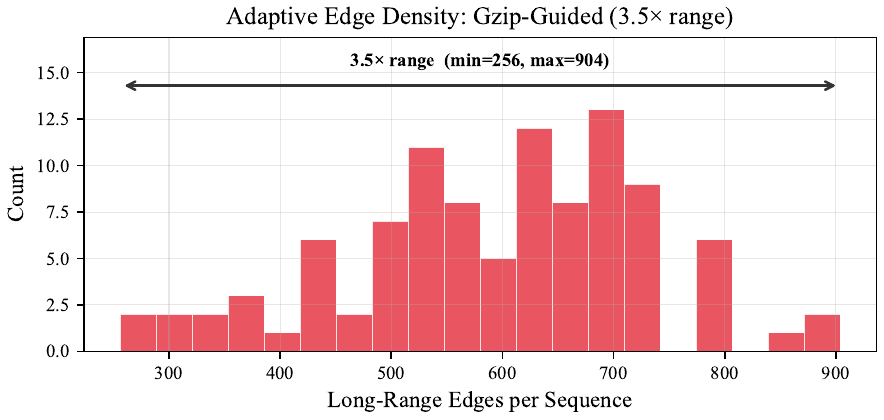}
\caption{Distribution of long-range attention edges per sequence in the
Gzip-Guided method (1000 training sequences from PG-19). The 3.5$\times$
range (256--904) reflects automatic density adaptation to content redundancy.}
\label{fig:edge_dist}
\end{figure}
\section{Extended Related Work}
\label{app:extended_related}

This appendix expands the abbreviated ``Compression in ML and long-context
evaluation'' paragraph in Section~\ref{sec:related} into the two full
threads it draws on.

\paragraph{Compression in machine learning.}
Compression has long been studied as a proxy for structure and
predictability in data. Jiang et al.~\cite{jiang2023gzip} demonstrated
that gzip-based Normalized Compression Distance (NCD) enables competitive
text classification without any learned parameters, sparking significant
interest in compression-based ML. Follow-up work~\cite{raff2024neuralncd}
showed that the relationship between compression rate and classification
accuracy is more nuanced than initially understood. Compression has also
been used for anomaly detection, clustering, and similarity measurement
across domains. Our work extends this line of research in a new direction:
rather than using compression for \emph{classification} (comparing
documents), we use it for \emph{attention routing} (identifying which
parts of a single document carry non-redundant content).
To our knowledge, this is the first use of classical compression for
attention mask construction.

\paragraph{Long-context evaluation.}
Evaluating long-context utilization is challenging.
LongPPL~\cite{fang2025longppl} proposes selective perplexity metrics
that measure per-token benefit from distant context.
The Long Range Arena~\cite{tay2021lra} provides classification benchmarks
at 1K--4K tokens, though recent work questions whether its tasks truly
require long-range reasoning. RULER and HELMET evaluate
retrieval and reasoning in instruction-tuned models.
For base language models without instruction tuning, loss-based
evaluation remains the most reliable metric. We adopt bits-per-byte (BPB)
on PG-19~\cite{rae2020compressive} as our primary metric and supplement it
with context-scaling analysis that measures how BPB improves as available
context grows (Table~\ref{tab:context_scaling}).

\section{Detailed Comparison to Learned Mask Methods}
\label{app:method_comp}

\begin{table}[!htb]
\centering
\caption{Comparison of adaptive mask construction methods.
Our approach is the only one requiring zero additional parameters,
no custom gradient estimators, and no specialized kernels.}
\label{tab:method_comparison}
\begin{tabular}{lcccc}
\toprule
\textbf{Method} & \textbf{Extra Params} & \textbf{Differentiable?} & \textbf{Custom Kernel?} & \textbf{Mask Cost} \\
\midrule
BigBird            & 0        & N/A (fixed)    & No  & $O(N)$ \\
Routing Trans.     & $O(dK)$  & K-means        & No  & $O(NK)$ \\
SBM-Trans.         & $O(HKd)$ & STE            & No  & $O(N^2K)$ \\
NSA                & $O(d)$   & Gating         & Yes & $O(N)$ \\
DMA                & $O(d^2)$ & End-to-end     & Yes & $O(Nd)$ \\
\textbf{Ours}      & \textbf{0} & \textbf{N/A (external)} & \textbf{No} & $O(N)$ \\
\bottomrule
\end{tabular}

\footnotesize{$d$ = model dim, $H$ = heads, $K$ = clusters, $N$ = seq length.}
\end{table}

Among the listed methods, our approach is the only one combining zero
extra parameters, no need for a gradient estimator through the mask
construction step, and no custom CUDA kernel. Routing Transformer and
SBM-Transformer add cluster-based parameters with associated optimization
challenges; NSA and DMA require specialized kernels for efficient
training.

\section{Convergence Speed Summary}
\label{app:convergence_table}

Table~\ref{tab:convergence} summarizes the training steps each method
requires to reach successive BPB thresholds, derived from the
trajectories in Table~\ref{tab:trajectories}.

\begin{table}[!htb]
\centering
\caption{\textbf{Convergence speed.} Training steps to reach BPB targets.
``---'' = not reached within 20K steps. Speedup is relative to BigBird
at the 2.5 BPB target.}
\label{tab:convergence}
\begin{tabular}{lcccc}
\toprule
\textbf{Method} & \textbf{3.0 BPB} & \textbf{2.5 BPB} & \textbf{2.0 BPB} & \textbf{Speedup} \\
\midrule
Longformer          & ---    & ---    & ---   & ---  \\
SBM-Transformer     & ---    & ---    & ---   & ---  \\
Dense               & $\sim$16K  & ---    & ---   & ---  \\
BigBird             & $\sim$9K   & $\sim$14K  & ---   & 1.0$\times$  \\
\midrule
Literal Pure        & $\sim$3.5K & $\sim$5.0K & $\sim$9.5K & 2.8$\times$ \\
\textbf{Gzip-Guided}& $\sim$\textbf{2.8K} & $\sim$\textbf{4.2K} & $\sim$\textbf{8.4K} & \textbf{3.3}$\times$ \\
\bottomrule
\end{tabular}
\end{table}

\section{Extended Discussion}
\label{app:extended_discussion}

This appendix continues the discussion begun in
Section~\ref{sec:discussion} with three further threads.

\paragraph{Why an external compression signal works.}
A natural question is why an externally-derived signal (gzip, no learned
parameters) outperforms a learned signal (SBM-Transformer, 247K parameters)
in our setting. Three factors stand out. First, gzip's signal is
\emph{noise-free}: every mask construction call returns a deterministic,
reproducible outcome. Learned masks, in contrast, must navigate
high-variance gradient estimates through Bernoulli sampling over $N^2$
potential edges. Second, gzip's signal is \emph{available prior to
training}: the structure of the input is captured before any model
weights are computed, providing stable supervision throughout training.
Learned mask parameters change as the model evolves, creating a
non-stationary mask distribution. Third, gzip is \emph{robust by
construction}: it has been engineered over decades to identify
within-document redundancy across diverse text, and natural-language
text plays to its strengths.

\paragraph{Implications for scaling.}
Our sequence-length scaling results (Table~\ref{tab:scaling}) show
the gap between fixed and compression-guided masks widening from
0.05 BPB at 4K to 0.63 BPB at 8K, with convergence speedup increasing
from 1.8$\times$ to 3.3$\times$. BigBird's delayed breakthrough shifts
later as context grows: step $\sim$4K at 4K, $\sim$7K at 8K. The
combinatorial explanation is straightforward: the number of possible
attention patterns over $B$ blocks grows as $2^{B^2}$, so fixed random
patterns become increasingly poor approximations of the useful routing
as $B$ grows. Compression-guided selection sidesteps the issue by
deterministically targeting non-redundant positions.

The failure of SBM-Transformer at 8K further supports this: learned
mask methods face their own scaling challenges (noisy gradients through
Bernoulli sampling over $N^2$ edges), while compression-based
construction scales trivially---gzip's $O(N)$ cost is independent of
the attention dimension and produces identical-quality signals at any
sequence length.

\paragraph{Beyond gzip.}
Our use of gzip is deliberately conservative: gzip is universally
available, fast, and well-understood. The same approach applies to any
general-purpose compressor: brotli, zstd, and LZ4 each offer different
trade-offs between compression strength and speed, and stronger
compressors may provide a sharper redundant/non-redundant signal.
Exploring whether stronger compression yields better attention
masks---and at what computational cost---is a natural direction for
future work.

\section{Detailed Limitations}
\label{app:limitations}

This appendix elaborates on the limitations summarized in
Section~\ref{sec:limitations}.

\paragraph{Scale.}
All experiments use a 92M-parameter model trained for 20K steps on a
single dataset (PG-19). While this is sufficient for controlled comparison
of attention mechanisms, it remains an open question whether the
advantages of compression-guided masks persist at billion-parameter scale,
longer training budgets, and diverse data mixtures. The information-theoretic
argument is scale-invariant in principle, but practical advantages may
diminish or strengthen at scale---this requires empirical validation.

\paragraph{Byte-level modeling.}
Our model operates on raw bytes (vocabulary size 256) rather than subword
tokens (vocabulary size 32K--128K). Byte-level models face a harder
sequence-length problem---8192 bytes corresponds to only $\approx$1500
words---but also have different attention dynamics than BPE-based models.
The compression ratios that guide our masks are computed on raw bytes;
adapting the method to subword-tokenized inputs would require computing
compression ratios at the token level, which is straightforward but
untested.

\paragraph{Gzip is not the strongest available compressor.}
Gzip at compression level 1 is fast but loose; stronger compressors
(brotli, zstd at high levels, neural compressors) would identify
within-block redundancy more accurately and may improve mask quality.
Conversely, our results show that even gzip at compression level 1
already produces highly effective masks, suggesting a degree of robustness
to compressor choice. Systematic comparison across compressors is left
to future work.

\paragraph{No true sparse computation.}
Our implementation materializes the full $N \times N$ attention matrix and
applies the sparse mask via element-wise multiplication. This provides no
wall-clock speedup over dense attention---the sparsity is ``logical'' rather
than computational. Achieving actual speedups requires integration with
block-sparse CUDA kernels (e.g., FlashAttention with block-sparse masks)
or custom sparse attention implementations, which we leave to future work.

\paragraph{SBM-Transformer comparison.}
Our SBM-Transformer baseline is a reimplementation, not the original
authors' code running on their original benchmark. While we followed the
official code faithfully (Section~\ref{sec:exp_sbm}), the SBM-Transformer
was designed for encoder-based classification on 1K--4K sequences (Long
Range Arena), not autoregressive language modeling at 8K. Its poor
performance in our setting may reflect a domain mismatch rather than a
fundamental limitation of learned masks. We encourage the community to
evaluate learned mask methods at longer contexts to determine whether
our finding generalizes.

\paragraph{Single dataset, no downstream evaluation.}
We evaluate only on PG-19 BPB. Production language models are
judged on downstream tasks (question answering, summarization, retrieval),
and it is possible that the representations learned through
compression-guided sparse attention, while achieving lower BPB, do not
transfer equally well to all downstream applications.

\section{Supplementary Figures}
\label{app:supp_figs}

This appendix collects visualizations referenced from the main text.
Each figure is a graphical complement to a table or paragraph already
presented above; the underlying numbers are unchanged.

\begin{figure}[!htb]
\centering
\includegraphics[width=0.85\textwidth]{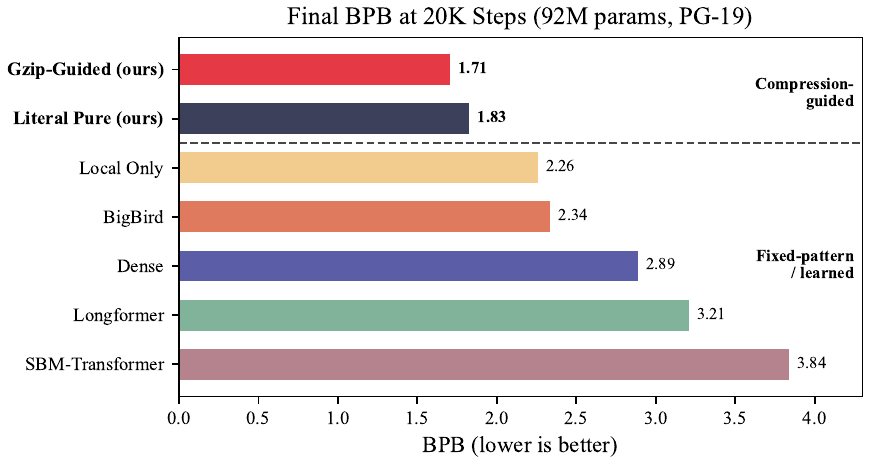}
\caption{\textbf{Visualization of Table~\ref{tab:main} (main results).}
Final BPB for all seven methods. Compression-guided methods
(bottom two) achieve 1.71--1.83 BPB, dramatically outperforming
fixed-pattern methods. SBM-Transformer, the only method with learned
mask parameters (+247K), performs worst.}
\label{fig:method_comparison}
\end{figure}

\begin{figure}[!htb]
\centering
\includegraphics[width=0.8\textwidth]{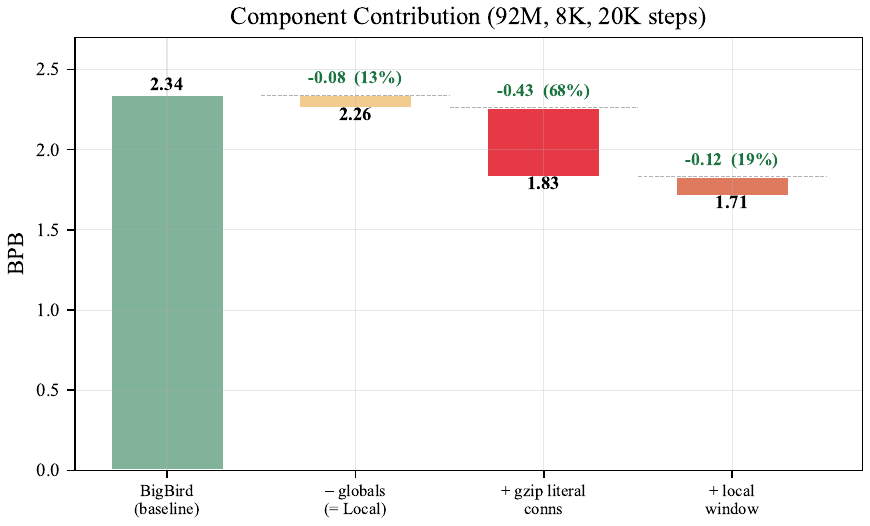}
\caption{\textbf{Visualization of Table~\ref{tab:ablation} (component
ablation).} Waterfall chart of component contributions.
Gzip-derived literal connections account for 68\% of the total
improvement over BigBird, the local window contributes 19\%, and
removing fixed global tokens contributes 13\%.}
\label{fig:ablation}
\end{figure}

\begin{figure}[!htb]
\centering
\includegraphics[width=\textwidth]{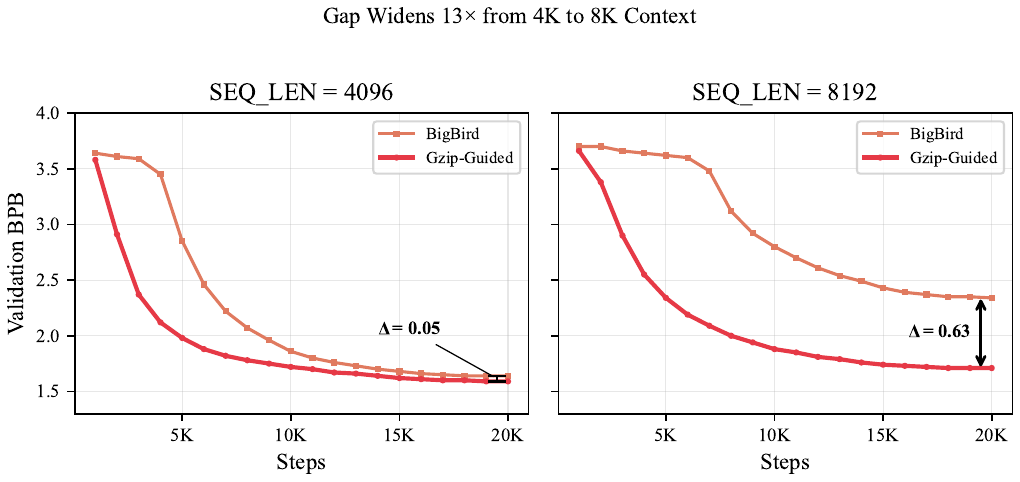}
\caption{\textbf{Visualization of Section~\ref{sec:exp_scaling}
(sequence length scaling).} Learning curves at 4K (left) and 8K (right)
context. At 4K, BigBird's delayed breakthrough occurs at step $\sim$4K
and converges within 0.05 BPB of Gzip-Guided. At 8K, the breakthrough
is delayed to step $\sim$7K with a final gap of 0.63 BPB.}
\label{fig:scaling_curves}
\end{figure}
\clearpage


\end{document}